\documentclass[sigconf,nonacm]{aamas}
\usepackage{amsmath}
\usepackage{tabularx}
\usepackage{multirow}
\usepackage{balance}
\usepackage{placeins}
\usepackage{needspace}
\usepackage{tikz}
\usetikzlibrary{arrows.meta,calc,backgrounds}
\title[Structure-Aware PPO-FP16]{Structure-Aware Numerical Stabilization for Half-Precision Proximal Policy Optimization}
\author{Linxuan Wu}
\affiliation{
  \institution{Institute of Automation, Chinese Academy of Sciences\\School of Artificial Intelligence, University of Chinese Academy of Sciences}
  \city{Beijing}
  \country{China}}

\author{Liang Chen}
\affiliation{
  \institution{Institute of Automation, Chinese Academy of Sciences\\School of Artificial Intelligence, University of Chinese Academy of Sciences}
  \city{Beijing}
  \country{China}}
\renewcommand{\shortauthors}{Linxuan Wu and Liang Chen}

\newcommand{\clip}{\operatorname{clip}}
\newcommand{\ours}{PPO-FP16}
\begin{document}
\pagestyle{plain}
\begin{abstract}
Local online training on resource-constrained platforms requires low computational and training-state overhead. We study the numerical stability and computational efficiency of half-precision proximal policy optimization (PPO) on commodity GPUs. Directly converting PPO training to FP16 is unreliable because numerical errors feed back through policy updates and subsequent data collection, while PPO repeatedly compares a current policy with a fixed behavior policy. We propose PPO-FP16 with three structure-aware modules. Global scalar protection stabilizes statistics shared across a minibatch or parameter update. Policy-change residual protection extracts and preserves small differences between the current and behavior policies. Efficient persistent-update protection stabilizes optimizer state and parameter writeback while reducing state storage. These changes preserve the PPO objective, hyperparameters, and training budget. Across six continuous-control tasks and five seeds, PPO-FP16 achieves the lowest mean return-AUC rank (1.90) among the four methods and ranks first in 17 of 30 environment--seed comparisons. Eight of nine performance configurations yield $1.16$--$2.03\times$ faster PPO updates. Relative to Full FP32 Adam, optimizer-state memory falls by $71.7$--$74.9\%$, and peak allocated GPU memory falls by up to $44.5\%$. The results show that protecting a small set of structurally important quantities can deliver stable, efficient FP16 PPO without retuning.
\end{abstract}
\maketitle
\thispagestyle{plain}
\section{Introduction}
\label{sec:introduction}

Reinforcement learning (RL) agents adapt their policies through continued interaction with an environment. This ability is valuable in robotics, autonomous systems, and control. An agent trained in simulation may need further training after deployment to accommodate differences in hardware, sensor noise, or task conditions. Local adaptation is especially useful when data cannot easily be uploaded, communication adds latency, or the device must operate independently for long periods. These resource-constrained settings motivate lower computational, memory, and energy costs for training~\cite{zhou2019edge}.

Low-precision training offers a direct way to reduce these costs. FP16 parameters and states use half the storage of FP32 values, and compatible hardware can provide higher arithmetic throughput~\cite{micikevicius2018mixed}. In continuous-control tasks, PPO often uses compact actor--critic networks, making it a promising candidate for deployment on resource-constrained edge devices. Yet half-precision PPO is difficult to stabilize. Each policy update changes subsequent actions and the data used for later updates. PPO also performs several optimization epochs on each rollout and uses a probability ratio between the current and behavior policies in its clipped objective~\cite{schulman2017ppo}. When the policy change is much smaller than either absolute log-probability, FP16 rounding can change the ratio and clipping decision. We study this numerical and computational problem on commodity GPUs as a step toward eventual use on resource-constrained platforms.

SAC-FP16 showed that algorithm-specific numerical reformulation can stabilize half-precision SAC training~\cite{bjorck2021lowprecision}. SAC and PPO, however, have different objectives and data flows. SAC uses replay and a maximum-entropy objective, whereas PPO collects an on-policy rollout and performs multiple updates relative to the fixed behavior policy that generated it. In PPO, errors in batch-level aggregate statistics, such as means, variances, and gradient norms, can affect an entire minibatch, small differences between the current and behavior policies determine clipping decisions, and gradients must pass through Adam states and the discrete FP16 parameter grid before producing persistent policy changes.

We therefore propose \ours{}. It preserves the PPO network architecture, objective, hyperparameters, and training budget while reorganizing three error-propagation paths. \textbf{Module A, global scalar protection,} stabilizes shared statistics through selective FP32 reductions and scale rearrangement. \textbf{Module B, policy-change residual protection,} extracts differences between Gaussian policies before forming absolute log-probabilities and performs clipping in the log domain. \textbf{Module C, efficient persistent-update protection,} builds on hAdam's root-second-moment representation. It introduces SHARP (Structured Half-precision Adam with Root-state Preconditioning), which shares second-moment state across parameter groups, and selectively compensates FP16 writeback of the log-standard-deviation. The residual-domain policy calculation and SHARP's structured state sharing are our main new designs.

Our contributions are:
\begin{itemize}
  \item We identify three sensitive error-propagation paths in FP16 PPO through stage-by-stage comparisons on fixed training data and numerical stress tests: shared statistics, small policy changes, and persistent parameter updates.
  \item We propose a structure-aware FP16 PPO design with three complementary protection modules while preserving the PPO objective and training configuration.
  \item We evaluate PPO-FP16 on six continuous-control tasks with five seeds. PPO-FP16 achieves the lowest mean return-AUC rank (1.90); eight of nine performance configurations yield $1.16$--$2.03\times$ faster PPO updates, optimizer-state memory is reduced by $71.7$--$74.9\%$ relative to Full FP32 Adam, and peak allocated GPU memory is reduced by up to $44.5\%$.
\end{itemize}

\section{Background}
\label{sec:background}

\subsection{Numerical Limits of Low-Precision Training}
IEEE FP32 has one sign bit, eight exponent bits, and 23 explicit fraction bits. FP16 has one sign bit, five exponent bits, and ten explicit fraction bits~\cite{ieee7542019}. Fewer exponent bits narrow the dynamic range, while fewer fraction bits widen the spacing between representable values. Training can consequently suffer from overflow, underflow, and loss of small terms during addition or subtraction. Early studies showed that successful limited-precision training depends on rounding, accumulation, scaling, and the placement of high-precision operations~\cite{gupta2015limited,wang2018eightbit,banner2018scalable}.

Recent methods control tensor scales and numerical dataflow more explicitly. Unit Scaling~\cite{blake2023unit} targets unit variance in weights, activations, and gradients, while Jetfire~\cite{xi2024jetfire} combines per-block quantization with an INT8 training dataflow. Recent low-precision studies have extended this direction to FP8 training, optimizer-state quantization, and long-horizon numerical stability~\cite{wortsman2023stable,peng2023fp8lm,xi2025coat,fishman2025scaling}. At lower precision, recent work studies how exponent--significand allocation affects training~\cite{sun2025scaling}, develops FP4 training recipes~\cite{wang2025fp4,cao2026metis}, and controls outliers or sensitive matrix multiplications through numerical and architectural transformations~\cite{ashkboos2025halo,hernandez2025fp8gemm}. These studies concern large language models rather than PPO, but reinforce the general principle that low-precision training depends on where and how numerical representations are applied.

\subsection{Low-Precision RL and SAC-FP16}
RL is commonly formulated as a Markov decision process $\mathcal M=(\mathcal S,\mathcal A,P,r,\gamma)$, where $\mathcal S$ and $\mathcal A$ are state and action spaces, $P(s'\mid s,a)$ is the transition probability, $r(s,a)$ is the reward, and $\gamma\in(0,1)$ is the discount factor. A policy $\pi_\theta(a\mid s)$, parameterized by $\theta$, maximizes expected discounted reward over trajectories it generates.
Because the policy also determines the training-data distribution, numerical errors can persist through a cycle of parameter updates, changed actions, new data, and further updates.

SAC-FP16~\cite{bjorck2021lowprecision} addresses low-precision training of soft actor-critic (SAC) with a stable root second moment, stable probability calculations, compensated updates, and compound scaling. Its hAdam component stores a root-second-moment state $w_t\approx\sqrt{v_t}$ instead of directly storing Adam's second moment $v_t$. This reduces the risk that small squared gradients underflow in FP16. SAC-FP16 demonstrates that numerical reformulation can stabilize \emph{SAC} in half precision.

SAC~\cite{haarnoja2018sac} and PPO~\cite{schulman2017ppo} differ in both objective and data flow. PPO has neither a target network nor the Jacobian term of a tanh-squashed SAC policy. Instead, it reduces advantages, clips a ratio against an old policy, and repeatedly optimizes on a fixed rollout. We therefore examined whether SAC-FP16 components transfer to PPO. Our exploratory transfer study found limited, environment-dependent gains from these components. This motivated a closer analysis of PPO's own numerical paths.

Recent work has also studied low-precision reinforcement learning at the scale of reasoning language models. QuRL~\cite{li2026qurl} accelerates rollout generation with INT8 or FP8 actors and addresses mismatch and small quantized weight updates. QaRL~\cite{gu2026qarl} aligns training-side computation with quantized rollouts to reduce the mismatch between a low-precision rollout policy and a full-precision learner. These methods focus on quantized LLM rollout generation, whereas our study concerns native half-precision PPO learner updates and persistent training state in continuous control.

\subsection{Proximal Policy Optimization}
PPO~\cite{schulman2017ppo} is an on-policy policy-gradient method. Let the behavior policy $\pi_{\theta_0}$ collect a rollout. For a sampled state--action pair $(s_t,a_t)$, the probability ratio is
\begin{equation}
r_t(\theta)=\frac{\pi_\theta(a_t\mid s_t)}{\pi_{\theta_0}(a_t\mid s_t)}
=\exp\!\left(\log\pi_\theta(a_t\mid s_t)-\log\pi_{\theta_0}(a_t\mid s_t)\right).
\end{equation}
Given advantage estimate $\widehat A_t$, the clipped surrogate objective is
\begin{equation}
L^{\mathrm{CLIP}}(\theta)=\mathbb E_t\!\left[
\min\!\left(r_t(\theta)\widehat A_t,
\clip(r_t(\theta),1-\epsilon_{\mathrm{clip}},1+\epsilon_{\mathrm{clip}})\widehat A_t\right)\right].
\label{eq:background-ppo}
\end{equation}
Here $\epsilon_{\mathrm{clip}}>0$ is the PPO clipping threshold and $\clip(x,l,u)=\min(\max(x,l),u)$. Clipping supports stable learning, but its decision depends on an accurate probability ratio.

Advantage estimates are obtained using standard generalized advantage estimation (GAE)~\cite{schulman2016gae} with the value network $V_\phi$. GAE accumulates temporal-difference residuals along the rollout, with $\lambda\in[0,1]$ controlling the bias--variance tradeoff.
The rollout is divided into minibatches and reused for several optimization epochs. GAE propagates information through time; log-probability differences enter an exponential; and repeated updates gradually separate the current and behavior policies. These connected operations allow local rounding errors to affect later updates.

\section{Method: Structure-Aware Half-Precision PPO}
\label{sec:method}

We propose a structure-aware design for half-precision PPO. Low-precision errors concentrate in three paths with broad or persistent effects. First, errors in means, variances, and gradient norms affect an entire minibatch or parameter update. Second, rounding and cancellation obscure small changes between the current and behavior policies. Third, Adam state updates and FP16 parameter writeback can distort the second moment or discard nonzero updates.

By \textbf{structure-aware}, we mean choosing numerical representations and computations according to a quantity's role in PPO and the reach of its errors. We distinguish global statistics shared across samples or parameters, policy differences that determine clipping, and optimizer and parameter states that persist across updates. We protect information at these propagation points.

\begin{figure*}[tp]
    \centering
    \includegraphics[width=\textwidth]{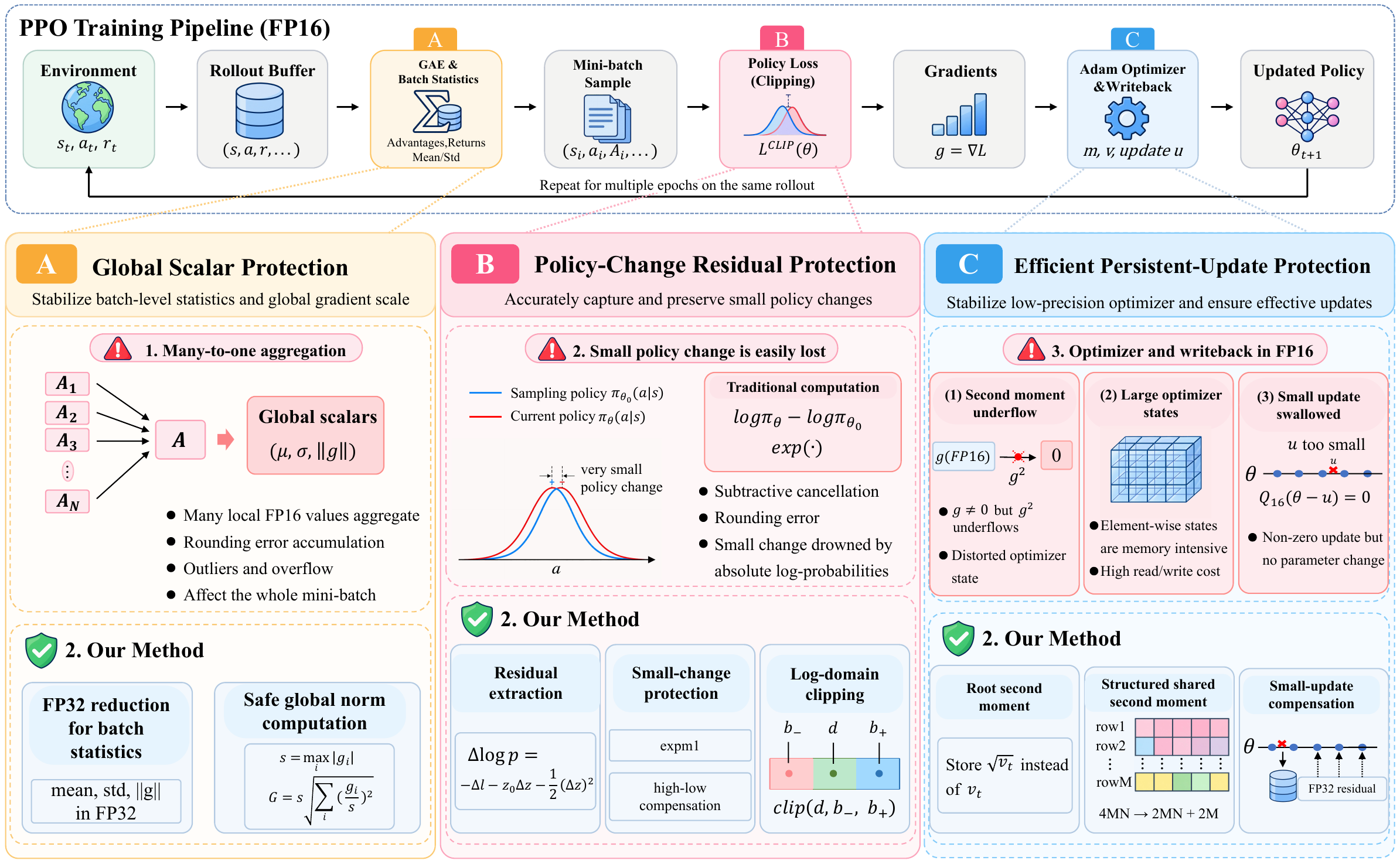}
    \Description{PPO-FP16 training pipeline and the three protection modules for global scalars, policy-change residuals, and persistent parameter updates.}
    \caption{PPO-FP16 training pipeline and its three structure-aware protection modules. Module A protects global statistics and gradient scale, Module B protects small policy changes in the residual domain, and Module C stabilizes optimizer state.}
    \label{fig:method_overview}
\end{figure*}

Figure~\ref{fig:method_overview} locates the three modules in the PPO update.

Parameters, the main forward and backward passes, gradients, and major optimizer states use FP16. FP32 is reserved for a few points where errors propagate across samples, time, or updates. By element count, FP32 accounts for less than $0.1\%$ of persistent training state. We write $Q_{16}(x)$ for rounding $x$ to FP16. Because GAE is recursive, each parallel environment maintains one FP32 recurrence state; rewards, values, advantages, and returns in the rollout remain FP16. The completed recurrence is written back to FP16 without retaining an FP32 advantage buffer.

\subsection{Module A: Global Scalar Protection}
\label{sec:module_a}

Our sensitivity analysis found that some local low-precision errors spread through global statistics. PPO includes several many-to-one computations: sums, means, variances, or norms reduce many sample or parameter values to a few statistics, which are then shared by a minibatch or parameter update. Advantage statistics, value-loss reduction, and the global gradient norm all have this form.

The advantage mean and variance normalize a whole minibatch, the batch-mean value loss sets the scale of critic gradients, and the global gradient norm determines one clipping coefficient shared by all parameters. Errors in these quantities therefore propagate beyond the values that first produced them.

For example, a minibatch of advantages $A_1,\ldots,A_N$ is normalized by subtracting its mean $\mu_A$ and dividing by its standard deviation $\sigma_A$, with a small stabilizer $\epsilon_A>0$ in the denominator. Both statistics are computed from the whole minibatch and then reused for every sample. The stabilizer $\epsilon_A$ is distinct from PPO's clipping threshold.

Because $\mu_A$ and $\sigma_A$ affect the whole minibatch, FP16 reduction error changes the scale of its policy loss and gradient. We therefore perform advantage statistics and value-loss reductions in FP32 while retaining FP16 data and main computation. This avoids repeated FP16 rounding of partial sums. Gradient norms have an additional failure mode before reduction.

For gradients $g_1,\ldots,g_M$, the global $L_2$ norm is $G=\sqrt{\sum_{i=1}^{M}g_i^2}$. PPO uses $G$ to determine a common gradient-clipping coefficient. Even when $g_i$ is representable in FP16, its square may exceed the largest finite FP16 value (about $6.55\times10^4$) before the sum begins.

We follow the classical scaled-norm approach~\cite{blue1978norm}. Let
\begin{equation}
    s = \max_i |g_i|,
    \qquad
    G = s\sqrt{\sum_{i=1}^{M}\left(\frac{g_i}{s}\right)^2}.
    \label{eq:scaled_gradient_norm}
\end{equation}
If $s=0$, we set $G=0$; otherwise Eq.~\eqref{eq:scaled_gradient_norm} is algebraically equivalent to the original norm, while $|g_i/s|\leq1$ prevents any squared term from exceeding one.

\subsection{Module B: Policy-Change Residual Protection}
\label{sec:module_b}

PPO performs multiple optimization epochs on data from a fixed behavior policy. Clipping constrains each update, so the current and behavior policies are often close.

PPO measures policy change through the probability ratio defined in Sec.~\ref{sec:background}. Here $\pi_{\theta_0}$ collected the rollout and $\pi_{\theta}$ is being optimized. For a state--action pair, write this ratio as $r$. Computation usually avoids direct division of probabilities by using the log ratio $d=\log\pi_\theta(a\mid s)-\log\pi_{\theta_0}(a\mid s)$, so $r=e^d$. For nearby policies, clipping depends on a near-zero $d$, rather than either absolute log-probability.

Even when a small $d$ is representable, the usual FP16 path first forms two absolute log-probabilities and subtracts them. Rounding in both absolute values can obscure their much smaller difference. We instead extract policy-parameter differences early and use \texttt{expm1}, which evaluates $e^x-1$ more accurately near zero, together with high--low compensated summation.

A separate FP16 exponential may round $e^x$ to one before subtracting one. Applying \texttt{expm1} directly to the small residual avoids this premature loss.

High--low compensated summation reduces further loss in local additions and subtractions. We use a TwoSum-form transformation~\cite{ogita2005accurate} to retain a rounded high-order component $h$ and a compensating low-order component $l$. We apply it to the terms of $\Delta z$ and the one-dimensional log-probability residual, then reduce high and low parts separately across action dimensions in FP32.

\Needspace{9\baselineskip}
For continuous actions, the policy $\pi_\theta(a\mid s)$ is a diagonal Gaussian density: $a\mid s\sim\mathcal{N}(\mu_\theta(s),\operatorname{diag}(\sigma_\theta^2(s)))$. We write $p_\theta(a\mid s)$ for this density and $\log p$ for its log-density; thus, $\log p\equiv\log\pi_\theta(a\mid s)$.
For one action dimension, $a\sim\mathcal{N}(\mu,\sigma^2)$. Writing $\ell=\log\sigma$ and $z=(a-\mu)e^{-\ell}$, its standard log-density is $\log p=-\ell-\tfrac12 z^2-C$, where $C=\tfrac12\log(2\pi)$. This representation makes the difference between nearby Gaussian policies explicit.

Rather than evaluating two complete log-probabilities, let $(\mu_0,\ell_0)$ be the behavior-policy parameters and $z_0=(a-\mu_0)e^{-\ell_0}$. Define the small residuals $\Delta\ell=\ell-\ell_0$, $\Delta\mu=\mu-\mu_0$, and $\Delta z=z-z_0$.

Since $z=z_0+\Delta z$, the log-probability difference is
\begin{equation}
    \Delta\log p
    =
    -\Delta\ell
    -
    z_0\Delta z
    -
    \frac{1}{2}
    (\Delta z)^2.
    \label{eq:residual_logprob}
\end{equation}

The common constant cancels without constructing either complete Gaussian log-probability.

To compute $\Delta z$, set $q=\Delta\mu\,e^{-\ell_0}$. Substituting the new mean and log-standard-deviation into $z=(a-\mu)e^{-\ell}$ gives $z=(z_0-q)e^{-\Delta\ell}$. Because nearby policies have small $\Delta\ell$, define $t=\operatorname{expm1}(-\Delta\ell)=e^{-\Delta\ell}-1$. Subtracting $z_0$ now yields
\begin{equation}
    \Delta z
    =
    z-z_0
    =
    z_0t
    -
    q(1+t).
    \label{eq:delta_z}
\end{equation}

This expresses the unchanged probability ratio through small residuals, allowing \texttt{expm1} and compensation to act before rounding and cancellation accumulate.

Limited-precision rollout storage can also make the reconstructed behavior log-probability differ from its sampling-time value. Without correction, PPO would measure change relative to this reconstructed policy rather than the policy that generated the rollout.

We therefore store a sampling-reference residual. At sampling time, the policy mean and log-standard-deviation are converted to FP32 to construct the Gaussian; the sampled action yields an FP32 log-probability $L_0$, temporarily held in an FP32 buffer. After the rollout, we reconstruct $\widetilde{L}_0$ in FP32 from the saved low-precision action and policy parameters. Their difference is rounded once when written to the rollout buffer:
\begin{equation}
    c
    =
    Q_{16}
    \left(
        L_0-\widetilde{L}_0
    \right).
    \label{eq:sampling_residual}
\end{equation}

During optimization, $c$ restores the sampling-time reference without a full high-precision rollout.

Clipping can be decided before forming $r=e^d$: $e^d>1+\epsilon_{\mathrm{clip}}$ exactly when $d>\log(1+\epsilon_{\mathrm{clip}})$, and analogously for the lower boundary. Define $b_+=\log(1+\epsilon_{\mathrm{clip}})$ and $b_-=\log(1-\epsilon_{\mathrm{clip}})$.

The advantage sign determines which side of the ratio is clipped. We can therefore write
\begin{equation}
    \bar d
    =
    \begin{cases}
        \min(d,b_+),
        & \hat A \geq 0,
        \\[4pt]
        \max(d,b_-),
        & \hat A < 0.
    \end{cases}
    \label{eq:log_domain_clip}
\end{equation}

This avoids an exponential intermediate in the clipping decision while preserving the PPO objective.

Module B therefore combines residual-domain Gaussian computation, \texttt{expm1}, compensated summation, sampling-reference correction, and log-domain clipping to preserve small policy changes without changing the clipped objective.

\subsection{Module C: Efficient Persistent-Update Protection}
\label{sec:module_c}

Adam~\cite{kingma2015adam} maintains persistent first- and second-moment states. The latter involves squared gradients. In FP16, a representable gradient may have a square too small to represent, distorting the second moment and potentially destabilizing training. Recent convergence analysis likewise identifies weights and second moments as sensitive components when adaptive optimizers are quantized~\cite{tang2026convergence}.

Building on hAdam from SAC-FP16~\cite{bjorck2021lowprecision}, we store $w_t=\sqrt{v_t}$ rather than the underflow-prone second moment $v_t$ and update it with a stable \texttt{hypot} operation. Our implementation retains Adam's first moment, bias correction, and update semantics; it changes the numerical representation of the second moment.

Standard Adam stores both moments for every parameter. For an $M\times N$ weight matrix, each state has $MN$ elements and must be read and written during updates. Low-precision optimizers compress these persistent states through 8-bit block-wise quantization~\cite{dettmers2022eightbit}, 4-bit moment quantization~\cite{li2023fourbit}, or FP8 representations~\cite{peng2023fp8lm,xi2025coat,fishman2025scaling}. SHARP instead reduces their granularity by sharing a root-second-moment scale within parameter groups.

Adafactor factorizes second moments~\cite{shazeer2018adafactor}; Shampoo and KrADagrad use structured matrix preconditioners~\cite{gupta2018shampoo,mei2023kradagrad}; and 4-bit Shampoo quantizes preconditioner state~\cite{wang2024fourbitshampoo}. SHARP instead shares only the root-second-moment scale within PPO parameter groups.

Building on hAdam, we introduce \textbf{SHARP} (Structured Half-precision Adam with Root-state Preconditioning), whose new contribution is structured state sharing for PPO parameters.

SHARP maintains the first-moment state element-wise, while the root-second-moment state is shared within predefined parameter groups. For an $M\times N$ hidden-layer weight matrix, we use row-wise grouping, yielding $M$ shared root-second-moment states instead of $MN$ element-wise second-moment states. Parameters within the same group therefore share only the preconditioning scale, while retaining independent first-moment states and parameter updates. Bias correction follows standard Adam, and the parameter update keeps the same Adam semantics; SHARP changes only the representation and granularity of the second-moment state. For small parameter groups and parameters that directly affect the policy output, we retain fine-grained states rather than enforcing sharing.

For each row, SHARP updates the shared root-second-moment from a scale-safe gradient root-mean-square through the stable \texttt{hypot} recurrence. Its ratio to the layer root-mean-square is clipped to $[0.25,4]$ before preconditioning.

With FP16 states, two element-wise Adam moments require $4MN$ bytes, whereas SHARP requires $2MN+2M$ bytes for an $M\times N$ row-grouped weight matrix. Thus, for large $N$, SHARP approaches a $50\%$ reduction relative to element-wise FP16 Adam states and a $75\%$ reduction relative to Full FP32 Adam states. Fewer second-moment values also reduce state reads and writes.

FP16 writeback can also swallow small updates. Let $\theta_t$ be a stored parameter and $u_t$ its optimizer update at step $t$. If $u_t$ is below the local FP16 spacing, $u_t\neq0$ may still produce
\begin{equation}
    Q_{16}(\theta_t-u_t)
    =
    \theta_t,
    \label{eq:update_swallowing}
\end{equation}
Repeated swallowed updates can separate the FP16 trajectory from the intended trajectory.

We add \textbf{small-update compensation}. In our implementation it applies only to the Gaussian policy's log-standard-deviation, which controls exploration scale. When a nonzero update does not change this FP16 parameter, we accumulate the unwritten change in an FP32 residual. Later accumulated updates can cross the FP16 grid spacing and be written to the parameter.

Module C thus protects both optimizer state and parameter writeback: SHARP reduces persistent second-moment storage and traffic, while selective residual writeback prevents repeated loss of small log-standard-deviation updates.

\subsection{Computational and Storage Complexity}
\label{sec:complexity}

Let $B$, $R$, $D$, $P$, and $G$ denote minibatch size, rollout samples, action dimensions, parameters, and second-moment groups. All modules preserve PPO's asymptotic update cost. Module A adds one $O(P)$ maximum reduction; Module B remains $O(BD)$ and adds $2R$ temporary bytes; Module C remains $O(P)$ but reduces the root-second-moment state from $P$ to $G\ll P$ elements. For an $M\times N$ row-grouped layer, optimizer state changes from $4MN$ to $2MN+2M$ bytes, approaching reductions of $50\%$ versus element-wise FP16 Adam and $75\%$ versus Full FP32 Adam. In practice, eight of nine configurations achieve $1.16$--$2.03\times$ update speedups, while optimizer-state memory falls by $71.7$--$74.9\%$ relative to Full FP32 Adam (Sec.~\ref{sec:efficiency}).

\section{Experiments}
\label{sec:experiments}

We evaluate numerical sensitivity, the contributions of Modules A--C, long-horizon learning quality, compute and memory costs, and the limits of lower-precision formats.

\Needspace{11\baselineskip}
\noindent\textbf{Implementation details.}
All long-horizon methods use the same two-layer $2\times64$ tanh actor and critic, PPO objective, and hyperparameters, with no environment- or precision-specific retuning. Runs use 2,048-step rollouts, 10 PPO epochs, 32 minibatches of 64 samples, and five seeds for 512,000 environment steps. The learning rate is $3\times10^{-4}$ with linear annealing, $\gamma=0.99$, GAE $\lambda=0.95$, the PPO clipping threshold is $0.2$, and the global gradient-clipping threshold is $0.5$. Adam uses $(\beta_1,\beta_2,\epsilon)=(0.9,0.999,10^{-5})$. Full FP32, AMP-FP16, AMP-BF16, and PPO-FP16 share these settings. PPO-FP16 keeps the main training state in FP16, with FP32 accounting for less than $0.1\%$ of persistent training-state elements. Experiments use PyTorch 2.3.0, CUDA 12.1, and an RTX 4060 Laptop GPU, which natively supports FP16 and BF16 computation. We use native PyTorch dtypes and autocast rather than fake quantization or explicit quantize--dequantize emulation. The Native-BF16 study uses actual BF16 tensors through a PyTorch reference implementation because the final SHARP CUDA kernel is FP16-specific. The nine-configuration performance benchmark uses one CUDA Graph for each complete PPO update.

\subsection{Sensitivity Analysis}
\label{sec:sensitivity}

We first converted the parameters, gradients, and Adam states of standard PPO directly to FP16. We compared this naive path against FP32 stage by stage on the same training data. Naive FP16 did not immediately produce NaNs or Infs, but iteration KL and clip fraction soon diverged from FP32, as did learning curves across seeds. Here iteration KL is the approximate policy divergence accumulated within an update iteration, and clip fraction is the fraction of minibatch samples whose probability ratios leave the unclipped interval. Numerical error accumulated through the PPO update cycle.

To locate the sources, we replayed the FP32 and FP16 paths on fixed rollouts and minibatches. We compared advantage statistics, losses, gradient norms, log-probabilities, log ratios, clipping decisions, Adam states, and actual parameter writeback. Stress tests covered extreme ranges rarely encountered in ordinary training. Holding the data fixed isolated differences in numerical representation and computation.

The first problem involved shared statistics. Errors in the advantage mean and variance altered the scale of a whole minibatch. The global gradient norm had a similar reach. Direct FP16 squaring and summation produced nonfinite norms for all 24 large-gradient stress inputs; scaled-norm computation produced finite norms for all 24. Reduction error and overflow during squaring both mattered.

The second problem involved the policy ratio. With Hopper states, actions, and old and new Gaussian parameters fixed, subtracting two absolute log-probabilities gave a mean absolute error (MAE) of $1.07\times10^{-3}$ in the local policy change. Constructing the change directly from policy-parameter differences reduced MAE to $4.7\times10^{-5}$. The absolute-log-probability path obscured the small difference that determines the ratio and clipping decision.

The third problem arose in optimizer state and writeback. Native FP16 Adam rounded $64.55\%$ of second-moment states to zero, and the update-magnitude ratio relative to FP32 Adam reached $124.07\times$. In the root-second-moment domain, the zero-state ratio fell to $0.58\%$. The update-magnitude ratio fell to $1.0722\times$. Later in training, many nonzero updates did not change FP16 parameters; accumulating them allowed eventual writeback. Thus both second-moment underflow and swallowed updates caused persistent error.

Together, these tests locate three connected paths: shared reductions spread local error, absolute probability calculations obscure small policy differences, and optimizer state and the FP16 grid retain error across updates. The three modules in Sec.~\ref{sec:method} address these paths.

\begin{figure*}[p]
\centering
\begin{minipage}[t]{.49\textwidth}
\centering
\captionof{table}{PPO update time and speedup over five processes.}
\label{tab:p93-performance}
\footnotesize
\setlength{\tabcolsep}{3.8pt}
\begin{tabular}{@{}rrrr@{}}
\toprule
Width / batch & FP32 (ms) & PPO-FP16 (ms) & Speedup \\
\midrule
32 / 512 & 0.307 & 0.213 & $1.44\times$ \\
32 / 1024 & 0.335 & 0.227 & $1.48\times$ \\
64 / 512 & 0.293 & 0.254 & $1.16\times$ \\
64 / 1024 & 0.319 & 0.260 & $1.23\times$ \\
512 / 1024 & 0.817 & 0.816 & $1.00\times$ \\
512 / 2048 & 1.308 & 0.963 & $1.36\times$ \\
1024 / 1024 & 2.410 & 1.639 & $1.47\times$ \\
1024 / 2048 & 4.145 & 2.318 & $1.79\times$ \\
2048 / 2048 & 15.837 & 7.784 & $2.03\times$ \\
\bottomrule
\end{tabular}
\end{minipage}\hfill
\begin{minipage}[t]{.49\textwidth}
\centering
\captionof{table}{Mixed-precision performance at batch size 2,048. Time is in ms; memory is in MiB.}
\label{tab:amp-performance}
\footnotesize
\setlength{\tabcolsep}{3.8pt}
\begin{tabular}{@{}lrrrr@{}}
\toprule
Method & Width & Time & Peak & State \\
\midrule
\multirow{3}{*}{Full FP32} & 64 & 0.363 & 16.69 & 0.085 \\
 & 512 & 1.385 & 27.09 & 4.176 \\
 & 2048 & 16.762 & 181.89 & 64.703 \\
\cmidrule(lr){1-5}
\multirow{3}{*}{AMP-FP16} & 64 & 0.395 & 16.69 & 0.085 \\
 & 512 & 1.050 & 27.09 & 4.176 \\
 & 2048 & 8.156 & 181.89 & 64.703 \\
\cmidrule(lr){1-5}
\multirow{3}{*}{AMP-BF16} & 64 & 0.401 & 16.69 & 0.085 \\
 & 512 & 1.039 & 27.09 & 4.176 \\
 & 2048 & \textbf{7.989} & 181.89 & 64.703 \\
\cmidrule(lr){1-5}
\multirow{3}{*}{PPO-FP16} & 64 & \textbf{0.291} & \textbf{16.54} & \textbf{0.023} \\
 & 512 & \textbf{1.019} & \textbf{21.82} & \textbf{1.058} \\
 & 2048 & 8.416 & \textbf{101.01} & \textbf{16.231} \\
\bottomrule
\end{tabular}
\end{minipage}

\vspace{5pt}
\includegraphics[width=.49\textwidth]{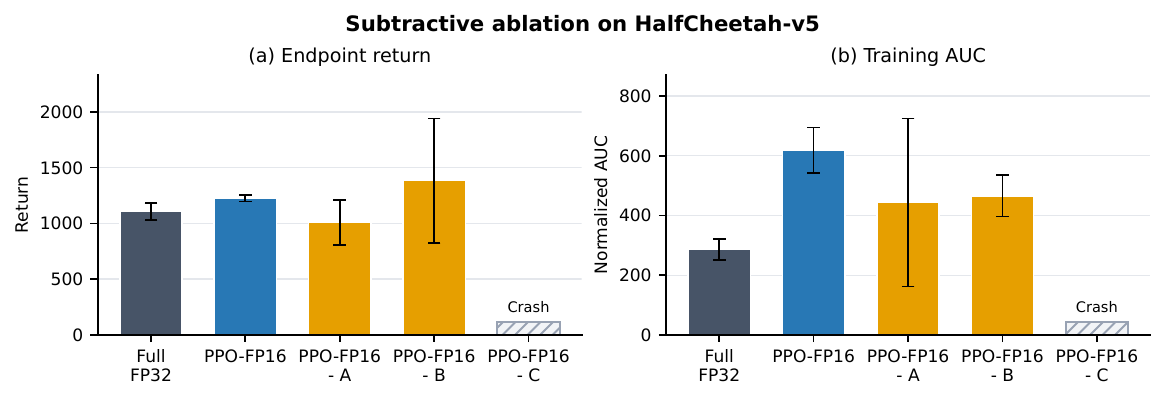}\hfill
\includegraphics[width=.49\textwidth]{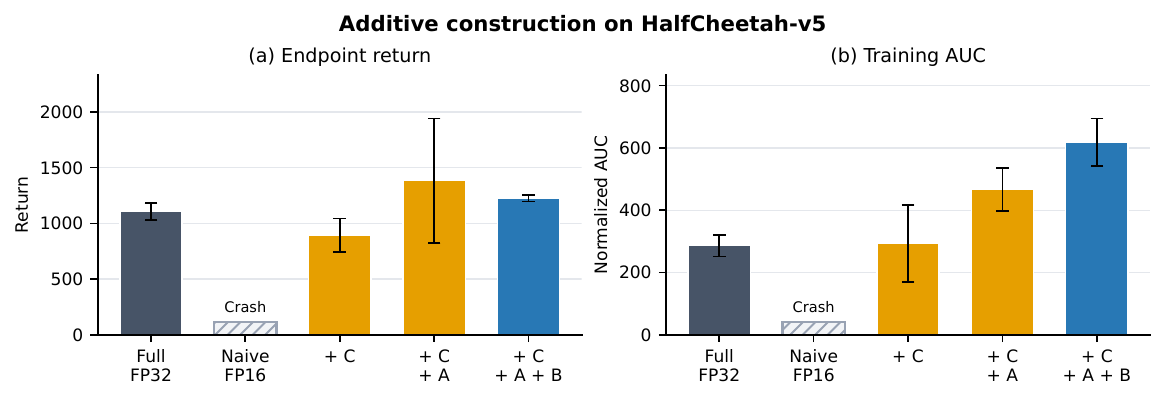}
\captionof{figure}{Module ablations on HalfCheetah-v5 over three seeds. Bars show mean $\pm$ standard deviation; Crash denotes instability.}
\label{fig:ablation-subtractive}
\label{fig:ablation-additive}

\vspace{3pt}
\includegraphics[width=.72\textwidth]{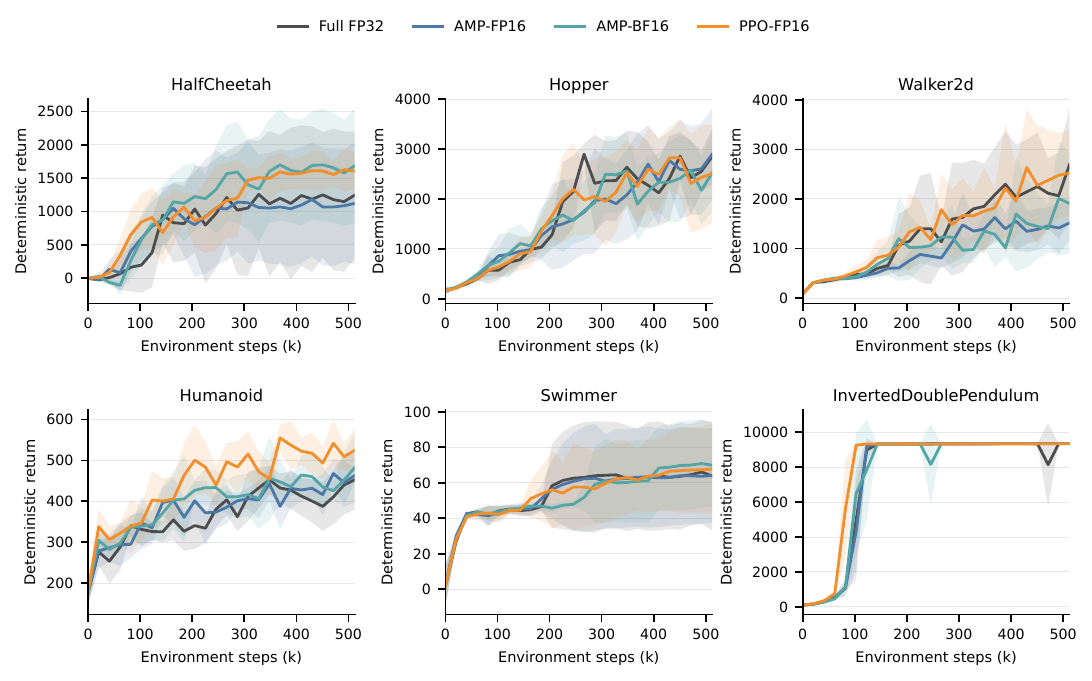}
\captionof{figure}{Long-horizon evaluation over six tasks and five seeds; shading shows 95\% confidence intervals.}
\label{fig:p96-quality}

\vspace{2pt}
\centering
\captionof{table}{Return AUC over five seeds (mean $\pm$ standard deviation). Bold marks the best mean per environment.}
\label{tab:p96-auc}
\footnotesize
\setlength{\tabcolsep}{5.5pt}
\begin{tabular}{@{}lrrrr@{}}
\toprule
Environment & Full FP32 & AMP-FP16 & AMP-BF16 & PPO-FP16 \\
\midrule
HalfCheetah & 832.4 $\pm$ 753.9 & 846.7 $\pm$ 672.2 & \textbf{1130.5 $\pm$ 594.6} & 1085.1 $\pm$ 258.5 \\
Hopper & \textbf{1684.1 $\pm$ 364.3} & 1608.2 $\pm$ 624.5 & 1609.1 $\pm$ 440.1 & 1667.0 $\pm$ 445.6 \\
Walker2d & 1321.3 $\pm$ 615.8 & 944.7 $\pm$ 183.2 & 1012.1 $\pm$ 331.3 & \textbf{1388.1 $\pm$ 443.9} \\
Humanoid & 366.8 $\pm$ 21.9 & 380.6 $\pm$ 12.5 & 397.1 $\pm$ 22.8 & \textbf{448.0 $\pm$ 33.3} \\
Swimmer & \textbf{54.4 $\pm$ 20.2} & 54.0 $\pm$ 18.7 & 52.9 $\pm$ 11.3 & 54.0 $\pm$ 15.3 \\
InvertedDoublePendulum & 7506.1 $\pm$ 241.7 & 7533.1 $\pm$ 139.2 & 7518.8 $\pm$ 270.3 & \textbf{7929.1 $\pm$ 125.3} \\
\midrule
Mean rank (30 pairs) & 2.83 & 2.63 & 2.63 & 1.90 \\
First-place count (30 pairs) & 7 & 5 & 1 & 17 \\
\bottomrule
\end{tabular}

\end{figure*}

\subsection{Module Ablations}
\label{sec:ablation}

We ablated the three modules in both subtractive and additive directions on HalfCheetah-v5, with fixed networks and hyperparameters. Each configuration used three seeds. We marked a configuration Crash if it failed a 20,480-step stability gate; the others were reinitialized and trained for 151,552 steps. Full FP32 served as a reference.

In the subtractive study, complete PPO-FP16 had the highest AUC and smallest endpoint variance. Removing A or B reduced AUC, while removing C caused a Crash. In the additive study, Naive FP16 crashed; adding C enabled training, and adding A and B raised AUC from 293.3 to 465.8 and 618.3. These results support complementary roles for the modules.

\subsection{Long-Horizon Learning Quality}
\label{sec:long-horizon}
\label{sec:quality}

We trained for 512,000 steps in six Gymnasium MuJoCo-v5 continuous-control environments~\cite{towers2024gymnasium,todorov2012mujoco}. We compared PPO-FP16 with Full FP32, AMP-FP16, and AMP-BF16. All methods used the same $2\times64$ tanh actor and critic, PPO objective, and hyperparameters. Each method ran five seeds without environment- or precision-specific retuning. Full FP32 used FP32 parameters, activations, gradients, and Adam states. AMP-FP16 kept FP32 parameters and Adam states while using FP16 autocast and GradScaler~\cite{micikevicius2018mixed}. AMP-BF16 likewise kept FP32 parameters and Adam states while using BF16 autocast~\cite{kalamkar2019bfloat16}. We define return AUC as the trapezoidal integral of deterministic evaluation return over environment steps, divided by the 512,000-step horizon. Figure~\ref{fig:p96-quality} shows deterministic evaluation curves, and Table~\ref{tab:p96-auc} reports return AUC.

To avoid bias from different return scales, we ranked the four methods by return AUC within each environment and seed, then averaged over 30 environment--seed combinations. PPO-FP16 had the best mean rank, 1.90, versus 2.63 for each AMP method and 2.83 for Full FP32. It ranked first in 17 comparisons, versus 7, 5, and 1 for Full FP32, AMP-FP16, and AMP-BF16, respectively.

PPO-FP16 had the highest mean AUC on Walker2d, Humanoid, and InvertedDoublePendulum, while maintaining stable learning on HalfCheetah, Hopper, and Swimmer. In paired comparisons, its AUC exceeded Full FP32, AMP-FP16, and AMP-BF16 in 20, 21, and 22 of the 30 environment--seed pairs, respectively.

For cross-environment comparison, we normalized each environment's AUC observations as $\widetilde A=(A-A_{\min})/(A_{\max}-A_{\min})$, where $A_{\min}$ and $A_{\max}$ are the minimum and maximum over all four methods and five seeds in that environment. We then used an environment-stratified paired bootstrap to calculate 95\% confidence intervals for method differences in $\widetilde A$. PPO-FP16's normalized AUC difference from AMP-FP16 was 0.208, with interval $[0.026, 0.371]$. The confidence intervals against Full FP32 and AMP-BF16 include zero, while PPO-FP16 achieves the lowest overall return-AUC rank and maintains strong long-horizon learning quality across all six tasks without FP32 master parameters or element-wise FP32 Adam states.
\subsection{Compute Performance and Memory}
\label{sec:efficiency}

Performance tests covered nine network-width and batch-size combinations on an RTX 4060 Laptop GPU. FP32 and PPO-FP16 performed the same complete PPO update. Each configuration used five independent processes, with 3,000 timed updates after warm-up.

PPO-FP16 achieved $1.16$--$2.03\times$ faster updates in eight of nine configurations; the remaining one was near parity with FP32. SHARP reduced optimizer-state memory by $71.7$--$74.9\%$ relative to Full FP32 Adam. Peak allocated GPU memory fell by up to $44.5\%$. Recent low-precision systems also demonstrate substantial training gains on consumer GPUs~\cite{schultheis2026llmq}; our measurements address a different regime, namely native FP16 PPO updates with compact actor--critic networks.

For a mixed-precision comparison, we tested widths 64, 512, and 2,048 at batch size 2,048 using Full FP32, AMP-FP16, AMP-BF16, and PPO-FP16. Each configuration used five paired processes, 500 warm-up iterations, and 1,000 timed updates. PPO-FP16 had the lowest update time at widths 64 and 512. At width 2,048 it was about $3$--$5\%$ slower than AMP, but reduced optimizer-state memory from 64.70 to 16.23 MiB and peak allocated GPU memory from 181.89 to 101.01 MiB. Tables~\ref{tab:p93-performance} and~\ref{tab:amp-performance} use independent benchmark protocols; absolute latency should be compared only within each table.

\subsection{Precision-Boundary Study}
\label{sec:other_formats}

We study exponent range and significand precision separately. BF16 retains an FP32-like exponent range but has fewer significand bits~\cite{kalamkar2019bfloat16}. Recent floating-point scaling studies also show that exponent allocation, significand length, and scaling granularity jointly determine low-precision training behavior~\cite{sun2025scaling}; FP4 methods therefore require additional scaling or structural treatment~\cite{wang2025fp4,cao2026metis}. In a \textbf{Native-BF16 variant}, we replace PPO-FP16 parameters, gradients, rollout data, first-moment state, and structured root-second-moment state with BF16 while retaining the three modules and their specified FP32 scalar boundaries. Because the final SHARP CUDA kernel targets FP16, we use an arithmetically equivalent PyTorch reference implementation. We compare learning quality, not BF16 speed, across the same six environments, five seeds, and 512,000 steps.

Across the six environments, Native-BF16 retained only $10.7$--$79.6\%$ of PPO-FP16's return AUC. A wider exponent range alone therefore did not preserve learning quality. The pattern is consistent with greater loss of small updates on BF16's coarser parameter grid, but direct causal attribution requires update-level measurements. Our residual writeback covers only the log-standard-deviation, so it cannot compensate writeback of the other network parameters.

We also held the exponent at five bits on HalfCheetah-v5 and shortened the main-tensor significand from E5M10 through E5M5. Each format ran two seeds for 100,352 steps. E5M9 still learned; E5M8 and E5M7 had lower returns and greater seed variation; E5M6 and E5M5 learned little. Together, these tests suggest that widening the exponent cannot replace the significand precision needed for persistent updates.

\section{Conclusion}
\label{sec:conclusion}

We presented PPO-FP16, a structure-aware approach that protects shared statistics, small policy changes, and persistent parameter updates while keeping parameters, gradients, and major optimizer states in FP16. Across six tasks and five seeds, it achieved the best mean return-AUC rank (1.90) and ranked first in 17 of 30 environment--seed comparisons. PPO updates were up to $2.03\times$ faster, optimizer-state memory was $71.7$--$74.9\%$ lower than with Full FP32 Adam, and peak allocated GPU memory was up to $44.5\%$ lower. BF16 and reduced-significand experiments delineated the remaining precision limits. These results establish the numerical and computational feasibility of FP16 PPO on commodity GPUs. Deployment on embedded accelerators, larger models, and more general low-precision methods remain future work.

\bibliographystyle{ACM-Reference-Format}
\bibliography{references}
\end{document}